\documentclass[conference]{IEEEtran}
\IEEEoverridecommandlockouts
\usepackage{cite}
\usepackage[pdftex]{graphicx}
\usepackage{amsmath}
\usepackage{algorithmic}
\usepackage{bm}
\usepackage{cleveref}
\usepackage{amssymb}
\usepackage{tikz}
\usepackage{adjustbox}
\usepackage{caption}
\usepackage[ruled]{algorithm2e}
\usepackage{subcaption}
\usepackage{booktabs}
\usepackage{float}
\usepackage{mathtools}
\usepackage{setspace}

\usepackage[skip=4pt]{caption}
\SetKwInput{KwInit}{Initialise}
\SetKwInput{KwOffline}{Offline}
\usetikzlibrary{arrows}
\begin{document}
\usetikzlibrary{calc}
\tikzstyle{decision} = [diamond, draw, fill=blue!20, 
    text width=6em, text badly centered, node distance=3cm, inner sep=0pt]
\tikzstyle{block} = [rectangle, draw, fill=blue!20, 
    text width=6em, text centered, rounded corners, minimum height=3.2em]
\tikzstyle{line} = [draw, -latex']
\tikzstyle{cloud} = [draw, ellipse,fill=red!20, node distance=3cm,
    minimum height=2em]
\newcommand{\coorTS}{\mathbf{\hat{o}}}  
\title{Anticipatory Navigation in Crowds by Probabilistic Prediction of Pedestrian Future Movements}

\author{Weiming Zhi$^{1}$, Tin Lai$^{1}$, Lionel Ott$^{1}$, Fabio Ramos$^{1,2}$
\thanks{{\tt\small firstName.lastName@sydney.edu.au}}%
\thanks{$^{1}$ School of Computer Science, The University of Sydney, Australia}%
\thanks{$^{2}$ NVIDIA, USA}
}

\maketitle

\begin{abstract}
Critical for the coexistence of humans and robots in dynamic
environments is the capability for agents to understand
each other's actions, and anticipate their movements. 
This paper presents Stochastic Process Anticipatory Navigation (SPAN), a framework that enables nonholonomic robots to navigate in environments with crowds, while anticipating and accounting for the motion patterns of pedestrians. To this end, we learn a predictive model to predict continuous-time stochastic processes to model future movement of pedestrians. Anticipated pedestrian positions are used to conduct chance constrained collision-checking, and are incorporated into a time-to-collision control problem. An occupancy map is also integrated to allow for probabilistic collision-checking with static obstacles. 
We demonstrate the capability of SPAN in crowded simulation environments, as well as with a real-world pedestrian dataset.
\end{abstract}

\section{Introduction}
Robots coexisting with humans often need to drive towards a goal while actively avoiding collision with pedestrians and obstacles present in the environment. To safely manoeuvre in crowded environments, humans have been found to anticipate the movement patterns in the environment and adjust accordingly~\cite{anticipate}. 
Likewise, we aim to endow robots with the ability to adopt a probabilistic view of their surroundings, and interact with likely future trajectories of pedestrians. Static obstacles in the environment are encoded in an occupancy map, and avoided during local navigation.
We present the Stochastic Process Anticipatory Navigation (SPAN) framework, which generates local motion trajectories that takes into account predictions of the anticipated movements of dynamics objects.

SPAN leverages a probabilistic predictive model to predict movements of pedestrians. Motion prediction for dynamic objects, in particular for human pedestrians, is an open problem and has typically been studied in isolation without direct consideration of its use in a broader planning framework. Many recent methods use learning-based approaches \cite{Alahi2016SocialLH,KTM,stmaps,socialGans} to generalise motion patterns from collected data. To adequately capture the uncertainty of these predictions, representations of future motion patterns of pedestrians need to be probabilistic. Furthermore, collision-checking with future pedestrian position needs to be done at a relatively high time resolution. To address these requirements, we represent uncertain future motions as continuous-time stochastic processes. Our representation captures prediction uncertainty and can be evaluated at an arbitrary time-resolution. We train a neural network to predict parameters of the stochastic process, conditional on the observed motion of pedestrians in the environment. Navigating through an environment requires online generation of collision-free and kinematically-feasible trajectories. We combine learned probabilistic anticipations of future motion and an occupancy map into a non-holonomic control problem formulation. To account for environment interactions a time-to-collision term \cite{Hayward1972NEARMISSDT} is integrated into the control problem. The formulated problem is non-smooth, and a derivative-free constrained optimiser is used to efficiently solve the problem to obtain control actions. A receding horizon approach is taken and the problem is re-optimised at fixed frequency to continuously adapt to pedestrian prediction updates.


SPAN is novel in jointly integrating data-driven probabilistic predictions of pedestrian motions and static obstacles in a control problem formulation. The technical contributions of this paper include: (1) A continuous stochastic process representation of pedestrian futures, that is compatible with neural networks, and allows for fast chance-constrained collision-checking, at flexible resolutions; 
(2) formulation of a control problem that utilises predicted stochastic processes as anticipated future pedestrian positions, to navigate through crowds. We demonstrate that the optimisation problem can be solved efficiently to allow the robot to navigate through crowded environments. SPAN also opens opportunities for advances of neural network-based motion prediction methods to be utilised for anticipatory navigation. The stochastic process pedestrian motion is compatible with many other neural network models as the output component of the network.    

\begin{figure}[t]
    \centering
    \includegraphics[width=0.45\textwidth]{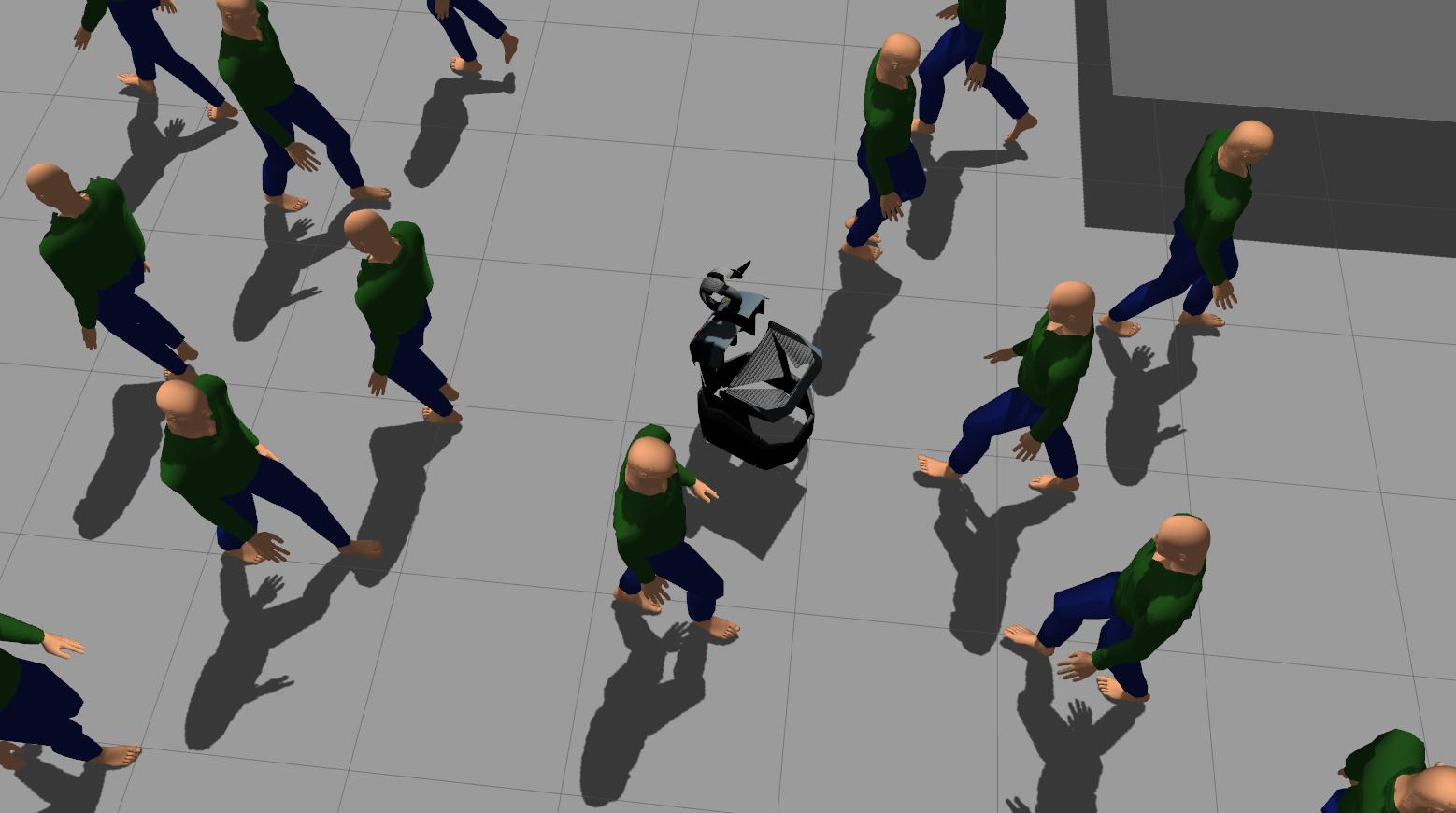}
    \caption{A non-holonomic robot drives towards a goal while interacting with pedestrians and static obstacles, and actively adapts to probabilistic predictions of surrounding movements.}
    \label{fig1}
\end{figure}


\section{Related Work}
SPAN actively predicts the motion patterns of pedestrians in the surrounding area and avoids collision while navigating towards a goal. Simple solutions to motion prediction include constant velocity or acceleration models \cite{humanSurvey}. Recent methods of motion prediction utilise machine learning to generalise motion patterns from observed data \cite{Alahi2016SocialLH,socialGans,KTM,stmaps}. These methods have a particular focus on capturing the uncertainty in the predicted motion, as well as on conditioning on a variety of environmental factors, such as the position of neighbouring agents. Generally, past literature examines the motion prediction in isolation, without addressing issues arising from integrating the predictions into robot navigation, where anticipation of pedestrian futures is valuable. 

We seek to generate collision-free motion, accounting for predicted motion patterns. Planning collision-free paths is well-studied, with methods at differing levels of locality. At one end of the spectrum, probabilistically-complete planning methods \cite{Lavalle98rapidly-exploringrandom, RRTtrees} explore the entire search-space to find a globally optimal solution. At the other end, local methods \cite{potential,rmpflow,CHOMP} aim to perturb collision-prone motion away from obstacles, trading-off global optimality guarantees for faster run-time. Our work focuses on the local aspect, where we aim to quickly obtain a sequence of control commands to navigate through dynamic environments. Various non-linear controllers \cite{NMPC,MPPI} also generate control sequences to avoid collision, but often assume obstacles are not time-varying. 

Local collision-avoidance methods for navigation for dynamic environments have been explored. Methods such as velocity obstacles (VO) \cite{Fiorini1998MotionPI} and optimal reciprocal collision avoidance (ORCA) \cite{Berg2009ReciprocalNC} find a set of feasible velocities for the robot, considering the current velocities of other agents. Partially observable Markov decision process (POMDP) solvers have been used in \cite{porca,hypdespot} to control linear acceleration for autonomous vehicles, but require steering controls from a global planner. Methods based on OCRA that indirectly solve for control sequences, such as \cite{GRCA}, are known to be overly conservative. To address this issue, an approach to directly optimise in the space of controls was proposed in \cite{NHTTC}, where a subgradient-based method utilises a time-to-collision value \cite{Hayward1972NEARMISSDT}, under constant velocity assumptions. Similarly, SPAN also optimises against time-to-collision, but using derivative-free constrained optimisation solvers. We extend the usage of time-to-collision to probabilistic predictions of pedestrian positions, provided by learning models, as well as integrating static obstacles captured by occupancy maps.   


\section{Problem Overview}
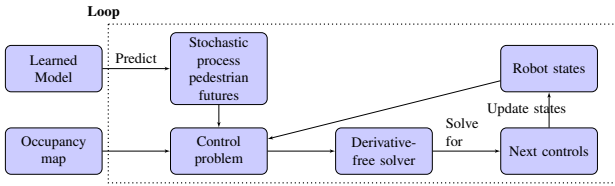
\begin{figure}[t]
\centering
\begin{adjustbox}{width=0.45\textwidth,center} 
\begin{tikzpicture}[node distance = 1.1cm]

    \node [block] (Model){Learned Model};
    \node [block, right of=Model, node distance=4cm] (Pred) {Stochastic process pedestrian futures};
    \node [block, below of=Model, node distance=2cm] (Map) {Occupancy map};
    \node [block, below of=Pred, node distance=2cm] (Control) {Control problem};
    \node [block, right of=Control, node distance=4cm] (solver) {Derivative-free solver};
    \node [block, right of=solver, node distance=4cm] (controls) {Next controls};
    \node [block, above of=controls, node distance=2cm] (States) {Robot states};
    \draw[black,thick,dotted] ($(Pred.north west)+(-1.5,0.2)$) node [text width=1cm,above]{\textbf{Loop}} rectangle ($(controls.south east)+(0.5,-0.2)$);

    \path [line] (Model) -- node [text width=1cm,midway,above]{Predict}(Pred);
    \path [line] (Map) -- (Control);
    \path [line] (Pred) -- (Control);
    \path [line] (Control) -- (solver);
    \path [line] (States) -- (Control);
    \path [line] (solver) -- node [text width=1cm,midway,above]{Solve for}(controls);
    \path [line] (controls) -- node [text width=3cm,midway]{Update states}(States);

\end{tikzpicture}
\end{adjustbox}
\caption{An overview of SPAN.}\label{Overview}
\end{figure}
A general overview of SPAN is provided in \cref{Overview}. (1) First, we train a model offline to predict probabilistic future pedestrian positions, conditioning on previous positions; (2) The predictive model and occupancy map are incorporate in an anticipatory control problem; (3) The control problem is solved online, in a receding horizon manner, with derivative-free solvers to obtain controls for the next states. This process is looped, with the learned model is continuously queried, and the problem continuously formulated and solved to obtain the next the next states.


\newcommand{\xCoor}{\bm{\hat{x}}}
\newcommand{\xState}{\mathbf{x}}

\subsection{Problem Formulation}
This paper addresses the problem of controlling a robot, to navigate towards a goal $\bm{g}$, in an environment with moving pedestrians and static obstacles.
The state of the robot at a given time $t$ , is given by $\xState(t)=\begin{smallmatrix}[\,x(t) & y(t) & \theta(t)\,]\end{smallmatrix}^{\top}$, denoting the robot's two dimensional spatial position and an additional angular orientation. We focus our investigation on wheeled robots, with the robot following velocity-controlled non-holonomic unicycle dynamics, where the system dynamics is given by:  
\begin{align}
    \dot{\xState}=f(\xState,\mathbf{u})=
    [v\cos(\theta),v\sin(\theta),\omega]^{\top},&& \mathbf{u}=[v,\omega]^{\top},\label{EqnMotion}
\end{align}
where $\dot{\xState}$ denotes the state derivatives. The robot is controlled via its linear velocity $v$ and angular velocity $\omega$. We denote the controls as $\mathbf{u}$ and the non-linear dynamics as $f(\xState,\mathbf{u})$. 

%
Information about \emph{static obstacles} in the environment is provided by an occupancy map, denoted as $f_{m}(\cdot)$, which represents the probability of being occupied $p(\mathrm{Occupied}|\xCoor)\in[0,1]$ at a coordinate $\xCoor\in \mathbb{R}^{2}$. The set of positions of $N$ pedestrian obstacles, at a given time $t$ from the current time, is denoted by the set $\mathcal{O}=\{\bm{o}^{1}(t),\ldots,\bm{o}^{N}(t)\}$. 
We are assumed to have movement data, allowing for training of a probabilistic model to predict future positions of pedestrians. 

\section{Probabilistic and Continuous Prediction of Pedestrian Futures}
In this section, we introduce a continuous \emph{stochastic process} representation of future pedestrian motion, and outline how the representation integrates into a neural network learning model. 
\subsection{Pedestrian Futures as Stochastic Processes}\label{representationSubsection}
Representations of future pedestrians positions need to capture uncertainty. Additionally, the time-resolution of the pedestrian position forecasts needs to synchronise with the frequency of collision-checks. These factors motivate the use of a continuous-time stochastic process (SP) for future pedestrian positions. A SP can be thought of as a distribution over functions. Furthermore, the continuous nature allows querying of future pedestrian position at an arbitrary time, rather than at fixed resolution, without additional on-line interpolation. 

We start by considering a deterministic pedestrian motion trajectory, before extending to a \emph{distribution over motion trajectories}. A trajectory is modelled by a continuous-time function, given as the weighted sum of $m$ basis functions. We write the $n^\text{th}$ pedestrian $\bm{o}^n\in\mathcal{O}$ coordinate at $t$ as:
\begin{align}
\mathbf{o}^n(t)={\mathbf{W}^n}^{\top}\bm{\Phi}(t), && \bm{\Phi}(t)=[\bm{\phi}(t,\mathbf{t}'),\bm{\phi}(t,\mathbf{t}')],\label{GPdef}
\end{align}
where ${\mathbf{W}^n}^{\top} \in \mathbb{R}^{m \times 2}$ is a matrix of weights that defines the trajectory. $\bm{\Phi}(t)$ contains two vectors of basis function evaluations, $\bm{\phi}(t,\mathbf{t}')$, one for each coordinate axis. $\bm{\phi}(t,\mathbf{t}')=[\phi(t,t'_{1}),\ldots,\phi(t,t'_{m})]^{\top}$ is a vector of basis function evaluations at $m$ evenly distanced time points, $\mathbf{t}'\in\mathbb{R}^{m}$. We use the squared exponential basis function given by $\phi(t,t')=\exp(-\gamma\|t-t'\|^{2})$, where $\gamma$ is a length-scale hyper-parameter.  

To extend our representation from a single trajectory to a distribution over future motion trajectories, we assume that the weight matrix is not deterministic, but random with a matrix normal ($MN$) distribution: 
\begin{align}
\mathbf{W}^{n}\sim MN(\mathbf{M}^{n},\mathbf{U}^{n},\mathbf{V}^{n}).
\end{align}
The matrix normal distribution is a generalisation of the normal distribution to matrix-valued random variables \cite{RandomMat}, parameters mean matrix $\mathbf{M}^{n}\in \mathbb{R}^{m\times 2}$ and scale matrices $\mathbf{U}^{n}\in \mathbb{R}^{m\times m}$ and $\mathbf{V}\in \mathbb{R}^{2\times 2}$. $\mathbf{M}^{n}$ is analogous to the mean of a normal distribution, and $\mathbf{U}^{n}, \mathbf{V}^{n}$ capture the covariance. A log-likelihood of the matrix normal distribution is given later in \cref{NLL}. Predicting a SP requires predicting $\mathbf{M}^{n},\mathbf{U}^{n},\mathbf{V}^{n}$.

\begin{figure}[t]
    \centering
    \includegraphics[width=0.3\textwidth]{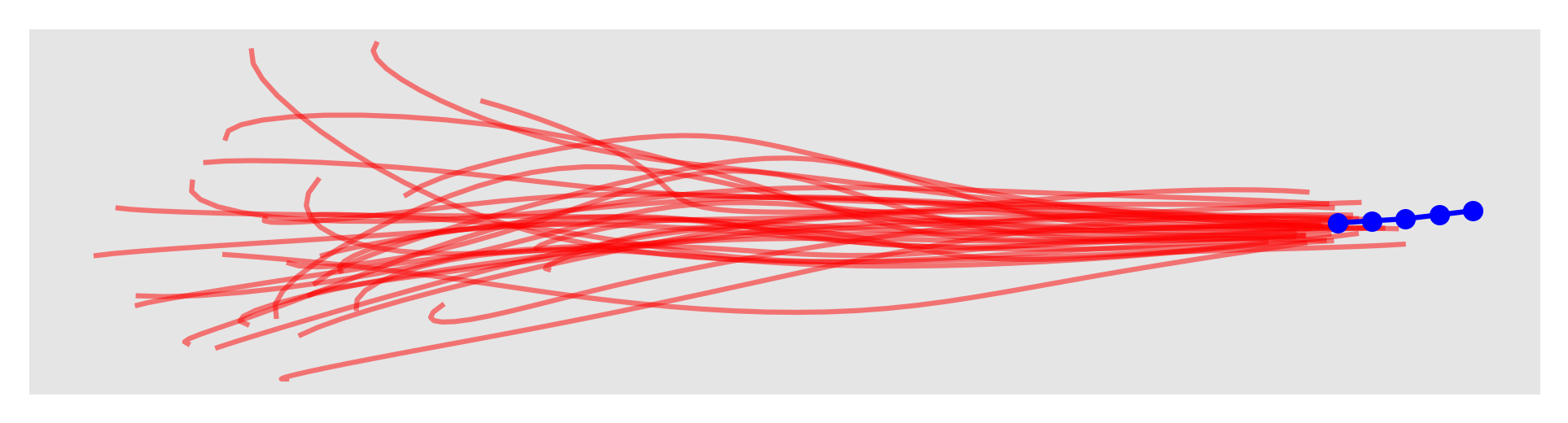}
    \caption{Sample trajectories (red) from a predicted distribution over future pedestrian motion, conditional on previously observed pedestrian coordinates (blue). }
    \label{pred_fig}
\end{figure}
\subsection{Learning Pedestrian Motion Stochastic Processes}\label{learn}
This subsection outlines how to train a model to predict SP representations of pedestrian motion. In the following subsection, we denote the weight matrix as $\mathbf{W}$ and parameters of the matrix normal distribution as $\mathbf{M},\mathbf{U},\mathbf{V}$, without referring to an obstacle index of a specific pedestrian. We aim to condition on a short sequence of recently observed pedestrian positions, up to the present $\{\coorTS_{1},\ldots,\coorTS_{p}\}$, and predict $\mathbf{M},\mathbf{U},\mathbf{V}$, which define a SP of the motion thereafter. A predictive model is trained offline with collected data, then queried online. 

\subsubsection{From timestamped data to continuous functions}
Pedestrian motion data is typically in the form of a discrete sequence of $n$ timestamped coordinates $\{t_{i},\coorTS_{i}\}_{i=1}^{n}$, where $t_{i}$ gives time and $\coorTS_{i}\in\mathbb{R}^{2}$ gives coordinates. Timestamped coordinates can be converted into a continuous function by finding a deterministic $\mathbf{W}$. For a pre-defined basis function $\bf{\Phi}(\cdot)$, we solve the least squares problem:  
\begin{equation}
{\min_{\mathbf{W}}\Big\{\sum_{i=1}^{n}(\coorTS_{i}-\underbrace{\mathbf{W}^{\top}\bm{\Phi}(t_{i})}_{\mathbf{o}(t_i)})^2+\lambda\|\mathrm{vec}(\mathbf{W})\|\Big\}}\label{lsuqares}
\end{equation}
where $\lambda$ is a regularisation hyper-parameter, and $\mathrm{vec}(\cdot)$ is the vectorise operator. By converting timestamped data to continuous functions, we can obtain a dataset, $\mathcal{D}$, with $N_{D}$ pairs, $\{(\{\coorTS_{1},\ldots,\coorTS_{p}\}_{d},\mathbf{W}_{d})\}_{d=1}^{N_{D}}$, where $\{\coorTS_{1},\ldots,\coorTS_{p}\}_{d}$ are previously observed coordinates, and $\mathbf{W}_{d}$ defines the continuous trajectory from $\coorTS_{p}$ onwards. $\mathbf{W}_{d}$ is obtained from timestamped coordinate data via \cref{lsuqares}. This dataset is used to train the predictive model.
\subsubsection{Training a Neural Network}
We wish to learn a mapping, $F$, from a sequence of observed pedestrian coordinates, $\{\coorTS_{1},\ldots,\coorTS_{p}\}$, to the parameters over distribution $\mathbf{W}$, that defines future motion beyond the sequence. We can use a neural network as,
\begin{equation}
    \overbrace{(\mathbf{M},\mathbf{U},\mathbf{V})}^{\mathclap{\text{defines SP over future motion}}}=\underbrace{F}_{\mathclap{\text{neural network function approximator}}}(\overbrace{\{\coorTS_{1},\ldots,\coorTS_{p}\}}^{\mathclap{\text{past coordinates}}})
\end{equation}
and train via the negative log-likelihood \cite{RandomMat} of the matrix normal distribution, over $N_{D}$ training samples of dataset $\mathcal{D}$. The loss function of the network is given as:
\begin{align}
    \mathcal{L}
    =\!-\!\sum_{d=1}^{N_{D}}\!\log\frac{\exp\{-\frac{1}{2}\mathrm{tr}[{\mathbf{V}}^{-1}\!(\mathbf{W}_d-\mathbf{M})^{\top}\!\mathbf{U}(\mathbf{W}_d-\mathbf{M})]\}}{(2\pi)^{m}|\mathbf{V}||\mathbf{U}|^{\frac{m}{2}}},\label{NLL}
\end{align}
where $\mathrm{tr}(\cdot)$ is the trace operator. In our experiments a simple 3 layer fully-connected neural network with width 100 units, and output layer giving vectorised $\mathbf{M},\mathbf{U},\mathbf{V}$ is sufficiently flexible to learn $F$. In this work, we focus on outlining a representation of motion able to be learned and used for collision-avoidance. More complex network architectures can be used to learn $F$, including recurrent network-based models \cite{Alahi2016SocialLH}.

\subsubsection{Querying the Predictive Model}
After training, the neural network outputs $\mathbf{M},\mathbf{U},\mathbf{V}$, conditional on observed sequences. $\mathbf{M},\mathbf{U},\mathbf{V}$ define random weight matrix $\mathbf{W}$, and the SP model via \cref{GPdef}. Individual trajectories can be realised from the SP by sampling $\mathbf{W}$. \Cref{pred_fig} shows sampled trajectories from a predicted SP.     

\section{Anticipatory Navigation with Probabilistic Predictions}
\subsection{Time-to-Collision Cost for Control}
This subsection introduces a control formulation which makes use of probabilistic anticipations of future pedestrian positions. Human interactions have been found to be governed by an inverse power-law relation with the time-to-collision \cite{powerlaw}, and can be integrated in robot control \cite{NHTTC}. To this end, we formulate an optimal control problem, with the inverse time-to-collision in the control cost:
\begin{subequations}
\begin{align}
    \min_{\mathbf{u}}\quad&\Big\{{\|\xCoor(T)-\bm{g}\|_{2}}+\frac{\kappa}{\tau(\xState_{0},\mathbf{u},\bm{o},f_{m})}\Big\}\label{costOpt}\\
    \textrm{s.t.}\quad&\dot{\xState}(t)=f(\xState(t),\mathbf{u}), \quad t\in[0,T], \label{costOpta}\\
    & \xState(0)=\xState_{0}, \label{costOptb}\\
    & \mathbf{u}_{lower}\leq\mathbf{u}\leq \mathbf{u}_{upper}\label{costOptc},
\end{align}
\end{subequations}
where $T$ is the time horizon; 
$\xCoor$, $\bm{g}$, $\xState$, $\mathbf{u}$ are the robot coordinates, goal coordinates, states, and target controls; $\tau$ is the time-to-collision value; $\kappa$ is a weight hyper-parameter controlling the collision-averseness; $\bm{o}$ denotes the position of pedestrians, $f_{m}$ denotes an occupancy map; $\mathbf{u}_{upper}, \mathbf{u}_{lower}$ denote the control limits, and $f$ refers to the system dynamics given in \cref{EqnMotion}. Like \cite{NHTTC}, we define time-to-collision as the first time a collision occurs under constant controls over the time horizon, enabling efficient optimisation.

\subsection{Chance-constrained Collision with Probabilistic Pedestrian Movements}
This subsection outlines a method to check for collisions between the robot and SP representations of pedestrian movement, to evaluate a time-to-collision value. Collision occurs when the distance between the robot and a pedestrian is below their collision radii. The set of coordinates where collision occurs, at time $t$, with the $n^{th}$ pedestrian is given by:
\begin{equation}
    \mathcal{C}^{n}_{t}=\left\{\bm{x}\in\mathbb{R}^2\;\middle|\;\|\bm{x}-\bm{o}^{n}(t)\|_{2}<r_{\xCoor}+r_{{o}_{n}}\right\},
\end{equation}
where $r_{\xCoor}$, $r_{\bm{o}_{n}}$ denote the radii of the robot and the pedestrian, and $\bm{o}^{n}(t)$ the pedestrian position. The position of pedestrians at a given time is uncertain, and given by the stochastic process defined in \cref{GPdef}. Therefore, we specify collision at $t$ in a chance constraint manner, where a collision between the robot and a pedestrian is assumed to occur if the probability of the robot position, $\xCoor(t)$, being in set $C^{n}_{t}$ is above threshold $\epsilon$, i.e. $p(\xCoor(t)\in \mathcal{C}^{n}_{t})>\epsilon$. Time-to-collision is taken as the first instance that a collision occurs, i.e., when the probability of collision with pedestrians exceeds $\epsilon$. Thus, for the chance-constraint formulation, the set of robot motions colliding at $t$ is given by, 
\begin{align}
    \mathcal{C}_{t}=\{\xCoor(t)| \max_{n}\{p(\xCoor(t)\in \mathcal{C}^{n}_{t})\}>\epsilon\}.\label{ttc}
\end{align}
As the weight parameters, $\mathcal{W}$, of the stochastic process is a matrix normal distribution, the position distribution at a specified time is a multivariate Gaussian. By considering the weight matrix distribution of the $n^{th}$ pedestrian, $\mathbf{W}^{n}$ and \cref{GPdef}, we have:
\begin{equation}
    \bm{o}^{n}(t)\sim \mathcal{N}({\mathbf{M}^{n}}^{\top}\bm{\Phi}(t),\mathbf{V}^{n}\bm{\Phi}^{\top}(t)\mathbf{U}^{n}\bm{\Phi}(t)),
\end{equation}
where $\mathbf{M}^{n}, \mathbf{U}^{n}, \mathbf{V}^{n}$ are the parameters of the distribution of weight matrix $\mathbf{W}^{n}$, and $\bm{\Phi}(t)$ contains basis function evaluations. The position of the robot $\xCoor(t)$ is deterministic, thus $\xCoor(t)-\bm{o}^{n}(t)$ is also multi-variate Gaussian, with a shifted mean. The probability of collision with the $n^{th}$ pedestrian is the integral of a Gaussian over a circle of radius $r_{\xCoor}+r_{\bm{o}_{n}}$. Let $\bm{d}^{n}=\xCoor(t)-\bm{o}^{n}(t)$, we have the expressions:
\begin{align}
    &p(\xCoor(t)\in \mathcal{C}^{n}_{t})=\int_{\|\bm{d}^{n}\|_{2}<r_{\xCoor}+r_{\bm{o}_{n}}}p(\bm{d}^{n})\mathrm{d}\bm{d}^{n}, \\
    &p(\bm{d}^{n})=\mathcal{N}(\xCoor(t)-{\mathbf{M}^{n}}^{\top}\bm{\Phi}(t),\mathbf{V}^{n}\bm{\Phi}^{\top}(t)\mathbf{U}^{n}\bm{\Phi}(t)).
\end{align}
This expression is intractable to evaluate analytically, and an upper bound to integrals of this kind is given in \cite{Zhu2019ChanceConstrainedCA} by:
\begin{equation}
    p(\xCoor(t)\in \mathcal{C}^{n}_{t})\leq \frac{1}{2}\bigg\{1+\mathrm{erf}\bigg(\frac{r_{\xCoor}+r_{\bm{o}_{n}}-\bm{a}^{\top}\bm{d}^{n}(t)}{\sqrt{2\bm{a}^{\top}\Sigma^{n}(t)\bm{a}}}\bigg)\bigg\},\label{upperbounds}
\end{equation}
where $\bm{a}=\frac{\bm{d}^{n}(t)}{\|\bm{d}^{n}(t)\|_{2}}$, $\Sigma^{n}(t) = \mathbf{V}^{n}\bm{\Phi}^{\top}(t)\mathbf{U}^{n}\bm{\Phi}(t)  \in \mathbb{R}^{2\times 2}$ is the covariance of the Gaussian $\bm{o}^{n}(t)$, and $\mathrm{erf}(\cdot)$ is the error function. Using the defined \cref{upperbounds}, we can efficiently evaluate the collision probability upper-bound between the robot and each pedestrian, and take the first time probability upper-bound exceeds $\epsilon$ as the time-to-collision with pedestrians. By \cref{ttc,upperbounds} a collision occurs at $t$ if:
\begin{equation}
    \max_{n=1,\ldots, N}\Bigg\{\frac{1}{2}\bigg[1+\mathrm{erf}\bigg(\frac{r_{\xCoor}+r_{\bm{o}_{n}}-\bm{a}^{\top}\bm{d}^{n}(t)}{\sqrt{2\bm{a}^{\top}\Sigma^{n}(t)\bm{a}}}\bigg)\bigg]\Bigg\}>\epsilon. \label{dynamicCheck}
\end{equation}
That is, collisions arise if the probability of collision with any pedestrian is above threshold $\epsilon$.
The robot-pedestrian time-to-collision, $\tau_{\bm{o}}$, is the earliest time collision with a pedestrian occurs, namely $\tau_{\bm{o}}=\min_{t\in[0,T]}(t)$, subject to \cref{dynamicCheck}.

\subsection{Chance-constrained Collision with Occupancy Maps}
Static obstacles in the environment are often captured via occupancy maps. Time-to-collision values accounting for occupancy maps can be integrated into SPAN. An occupancy map can be thought of as a function, $f_m(\xCoor)$, mapping from position, $\xCoor\in\mathbb{R}^{2}$, to the probability the position is occupied, $p(\mathrm{Occupied}|\xCoor)$. We ensure the probability of being occupied at any coordinate on the circle of radius, $r_{\xCoor}$, around our robot position is below the chance constraint threshold, $\epsilon$. By considering the occupancy map as an implicit surface, a collision occurs at time $t$ if,
\begin{align}
    \max_{\phi\in[-\pi,\pi)}\bigg\{f_m\bigg(\xCoor(t)+r_{\xCoor}\begin{bmatrix} \sin(\phi)\\ \cos(\phi) \end{bmatrix}\bigg)\bigg\}>\epsilon. \label{staticCheck}
\end{align}
Maximising the independent variable, $\phi$, sweeps a circle around the robot to check for collisions. As the domain of $\phi$ is relatively limited and querying from occupancy maps is highly efficient, taking under $10^{-6}$s each query, a brute force optimisation can be done. The time-to-collision due to static obstacles is then the first time-step where a collision occurs, namely $\tau_{m}=\min_{t\in[0,T]}(t)$, subject to \cref{staticCheck}. We take the overall time-to-collision value for the optimisation in \cref{costOpt} as the minimum of the time-to-collision for both static obstacles and pedestrians, specifically $\tau=\mathrm{minimum}(\tau_{\bm{o}},\tau_{m})$.

\subsection{Solving the Control Problem}
We take a receding horizon control approach, continuously solving the control problem defined in \cref{costOpt} - \cref{costOptc}, and updating the controls for a single time-step. Algorithm \ref{algorithm} outlines our framework, bringing together the predictive model, the formulation and solving of control problem. A predictive model of obstacle movements is trained offline with motion data. The predictive model and a static map of the environment are provided to evaluate the time-to-collision, $\tau$, in the control problem. We take a ``single-shooting'' approach and evaluate the integral in the control cost via an Euler scheme with step-size $\Delta t=0.1s$. At each time step we check for collision by \cref{dynamicCheck,staticCheck}, and $\tau$ is the first time-step a collision occurs. The control problem is non-smooth, so we use the derivative-free solver COBYLA \cite{Powell1994ADS} to solve the optimisation problem, and obtain controls $\mathbf{u}$. As the control-space is relatively small, the optimisation can be computed quickly. We can re-solve with several different initial solutions in the same iteration to refine and evade local minima. Controls are updated at $10$ Hz. 


\begin{algorithm}[tb]
\algsetup{linenosize=\footnotesize}
  \footnotesize
    \caption{\small{Stochastic Process Anticipatory Navigation}} \label{algorithm}
\Indmm\Indmm
    \KwIn{$\xState_0,\dot{\xState}_0,\mathbf{g},\mathcal{D},f_{m},\Delta t$}
    \KwOffline{Train neural network, $F$, to predict with dataset $\mathcal{D}$ (\cref{learn})}
     \KwInit{$t \gets 0$; $\xState(0)\gets \xState_0$; $\dot{\xState}(0)\gets \dot{\xState}_0$
        }
\Indpp\Indpp
    \While{goal not reach}{
        Observe $N$ pedestrians' coordinates, within a past time window, $\{\{\coorTS_{1},\ldots,\coorTS_{p}\}\}_{i=0}^{N}$\;
        $\mathcal{O}\gets\emptyset$ \tcp*{collects predictions}
        Predict $\{\mathbf{M}^n,\mathbf{U}^n,\mathbf{V}^n\}_{n=0}^{N}=F(\{\coorTS_{1},\ldots,\coorTS_{p}\}_{i})$;
        
        \For{$n=1$ to $N$\tcp{For all peds.}}{\tcp{Construct pedestrian futures as SP}
            $\bm{o}^n(t)={\mathbf{W}^n}^{\top}\bm{\Phi}(t)$, $\mathbf{W}^n\sim \mathrm{MN}(\mathbf{M}^n,\mathbf{U}^n,\mathbf{V}^n)$\;
            $\mathcal{O}\gets\mathcal{O}\cup \{\bm{o}^n(t)\}$\;
        }
        \tcp{optimise for control with pred.}
        $\mathbf{u} \gets$ Solve \cref{costOpt,costOpta,costOptb,costOptc}, with $\mathcal{O},f_{m},\xState(t)$ to get controls\; 
        $\xState(t+\Delta t)\gets\mathrm{NextState(\xState(t),\mathbf{u})}$ \tcp{Execute controls}
        $t \gets t+\Delta t$\;
        \If{$\|\xCoor(t)-\mathbf{g}\|_2 < \epsilon_{goal}$}{
            \texttt{Terminates}\tcp*{goal reached}
        }
    }
    
\end{algorithm}

\section{Experimental Evaluation}
We empirically evaluate the ability of SPAN to navigate in environments with pedestrians and static obstacles. Results and desirable emergent behaviour are discussed below. 
\subsection{Experimental Setup}
Both simulated pedestrian behaviour and real-world pedestrian data are considered in evaluations. We use a power-law crowd simulator \cite{powerlaw,karamouzas17}, and real-world indoor pedestrian data from \cite{thorDataset2019}. We construct two simulated environments, \emph{Sim 1} and \emph{Sim 2}, both with 24 pedestrians. \emph{Sim 1} models a crowded environment with noticeable congestion; \emph{Sim 2} models a relatively open environment with less frequent pedestrian interaction. The simulated pedestrians have a maximum speed of $1m/s$. Pedestrian trajectories collected at different times in dataset are overlaid, increasing the crowd density. The maps used in our experiments are shown in \cref{expref0}. We navigate the robot from starting point to a goal. Gazebo simulator \cite{gazebo} was used for visualisation. The following metrics are used:
\begin{enumerate}
    \item Time-to-goal (TTG): The time, in seconds, taken for the robot to reach the goal;
    \item Duration of collision (DOC): The total time, in seconds, the robot is in collision.  
\end{enumerate}
We evaluate SPAN against the following methods:
\begin{itemize}
\item Closed-loop Rapidly-exploring Random Trees$^*$ (CL-RRT$^*$). An asymptotic optimal version of \cite{kuwata2008motion} with a closed-loop prediction and a nonlinear pure-pursuit controller \cite{amidi1991integrated}. \cite{kuwata2008motion} demonstrates an ability to handle dynamic traffic, and is a DARPA challenge entry. At each iteration, a time budget of $0.1s$ is given to the planner to find a free trajectory towards the goal, with pedestrians represented as obstacles. If a feasible solution is not found, the robot will remain at its current pose. 
\item A reactive controller. The reactive controller continuously solves to find optimal collision-free controls for the next time-step, without anticipating future interaction.  
\end{itemize}
In our experiments, we set the time-to-collision weight, $\kappa$=100; the trajectory length-scale hyper-parameter, $\gamma=0.01$; the control limits on linear/ angular velocity as $-1\leq\bm{v}\leq 1$ and $0\leq \omega \leq 1$. Each iteration we solve for a ``look-ahead'' time horizon of $T=4.0s$. The collision threshold is set at $\epsilon = 0.25$, and the radii for the robot and the pedestrians are given as $r_{\xCoor},r_{\bm{o}} = 0.4m$. In the simulated setups, the predictive model is trained with additionally generated data from the crowd simulator; in the dataset setup, $80\%$ of the data was used to train the predictive model, with the remainder for evaluation. We condition on observed pedestrian movements in the last 0.5s, to predict the next 4.0s. Occupancy maps provided are implemented with the method in \cite{hilbertMap}. Controls were solved with different initialisations 40 times per iteration. The predictive model is implemented in Tensorflow \cite{tflow}, control formulation and COBYLA \cite{Powell1994ADS} are in FORTRAN, interfacing with Python2. Experiments run on a laptop with an Intel i7 CPU, and 32GB RAM. 
\subsection{Experimental Results}
\begin{figure}
\centering
     \begin{subfigure}{0.15\textwidth}
         \centering
         \frame{\includegraphics[width=\textwidth]{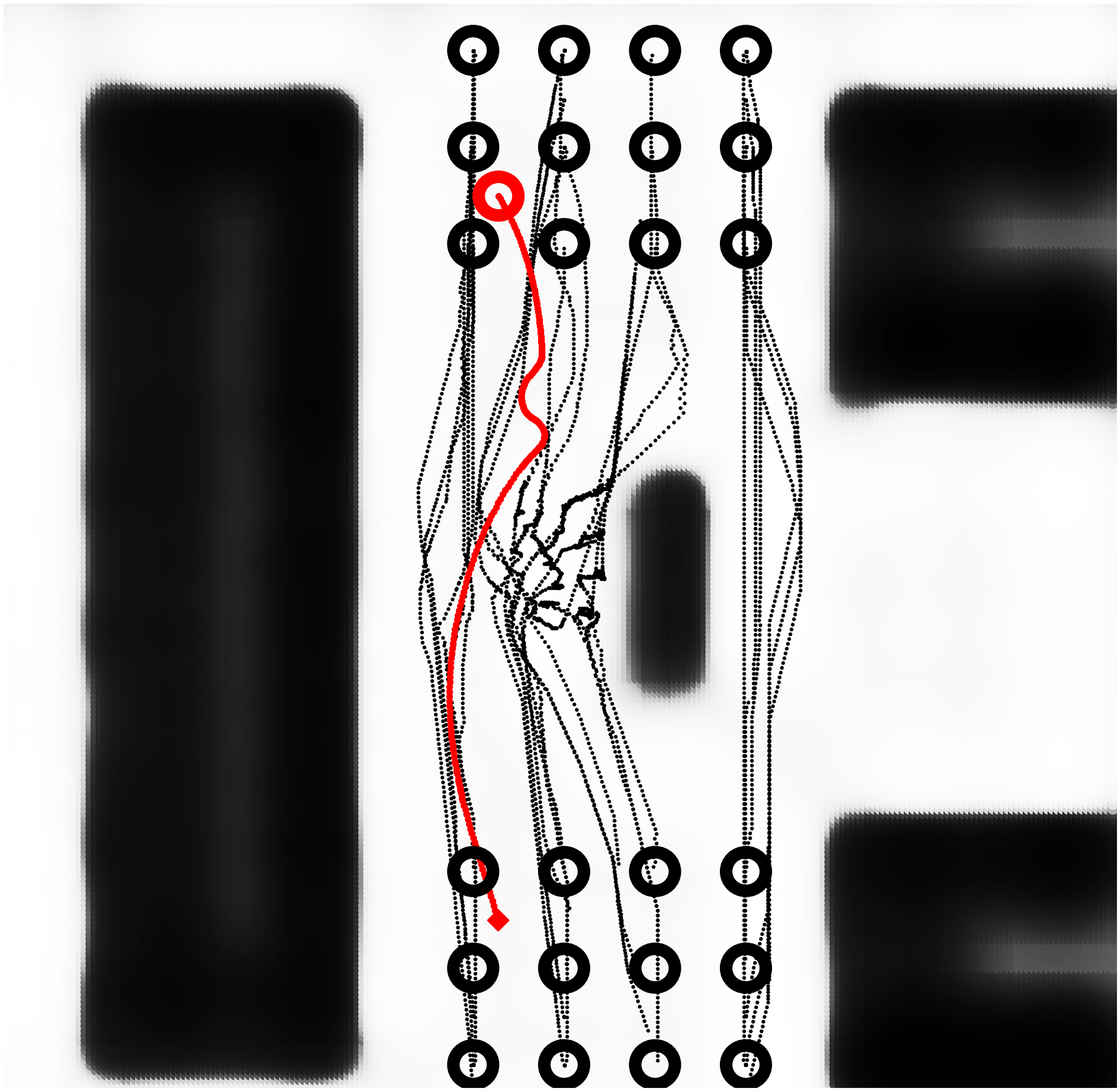}}
     \end{subfigure}
     \begin{subfigure}{0.15\textwidth}
         \centering
         \frame{\includegraphics[width=\textwidth]{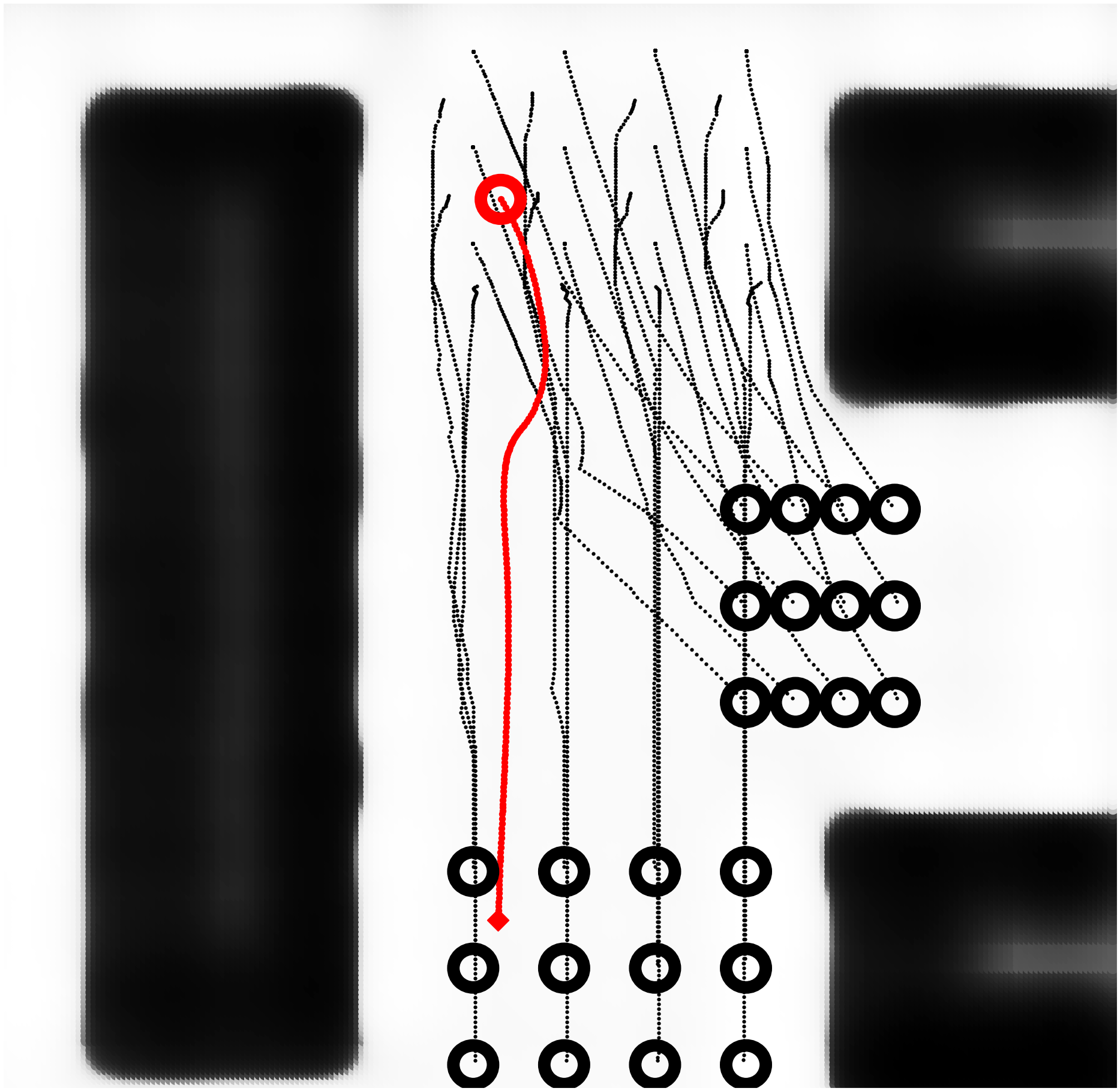}}
     \end{subfigure}
     \begin{subfigure}{0.15\textwidth}
         \centering
         \frame{\includegraphics[width=\textwidth]{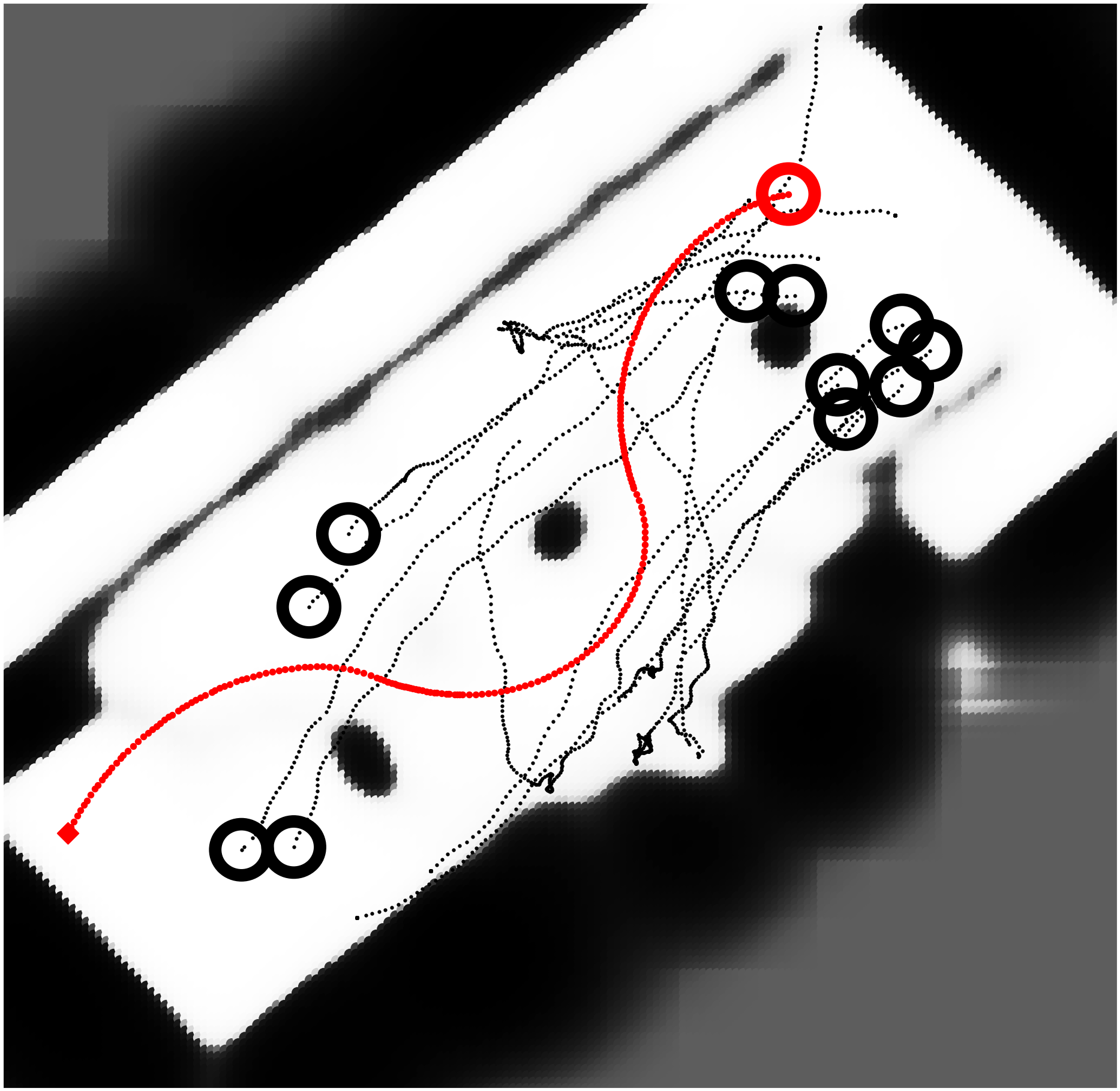}}
     \end{subfigure}
        \caption{Motion of pedestrians (black) and robot (red) overlaid on an occupancy map of the environments for Sim1 (left), Sim2 (center), and Dataset (right). Note that crossing trajectories does not necessarily indicate a collision, as moving agents could be at the same position at different times. The starting position of the pedestrians and robot are indicated by circles. The robot navigates smoothly and sensibly, remaining collision-free.}
        \label{expref0}
\end{figure}

\begin{figure}
\centering
     \begin{subfigure}{0.24\textwidth}
         \centering
         \frame{\includegraphics[width=\textwidth]{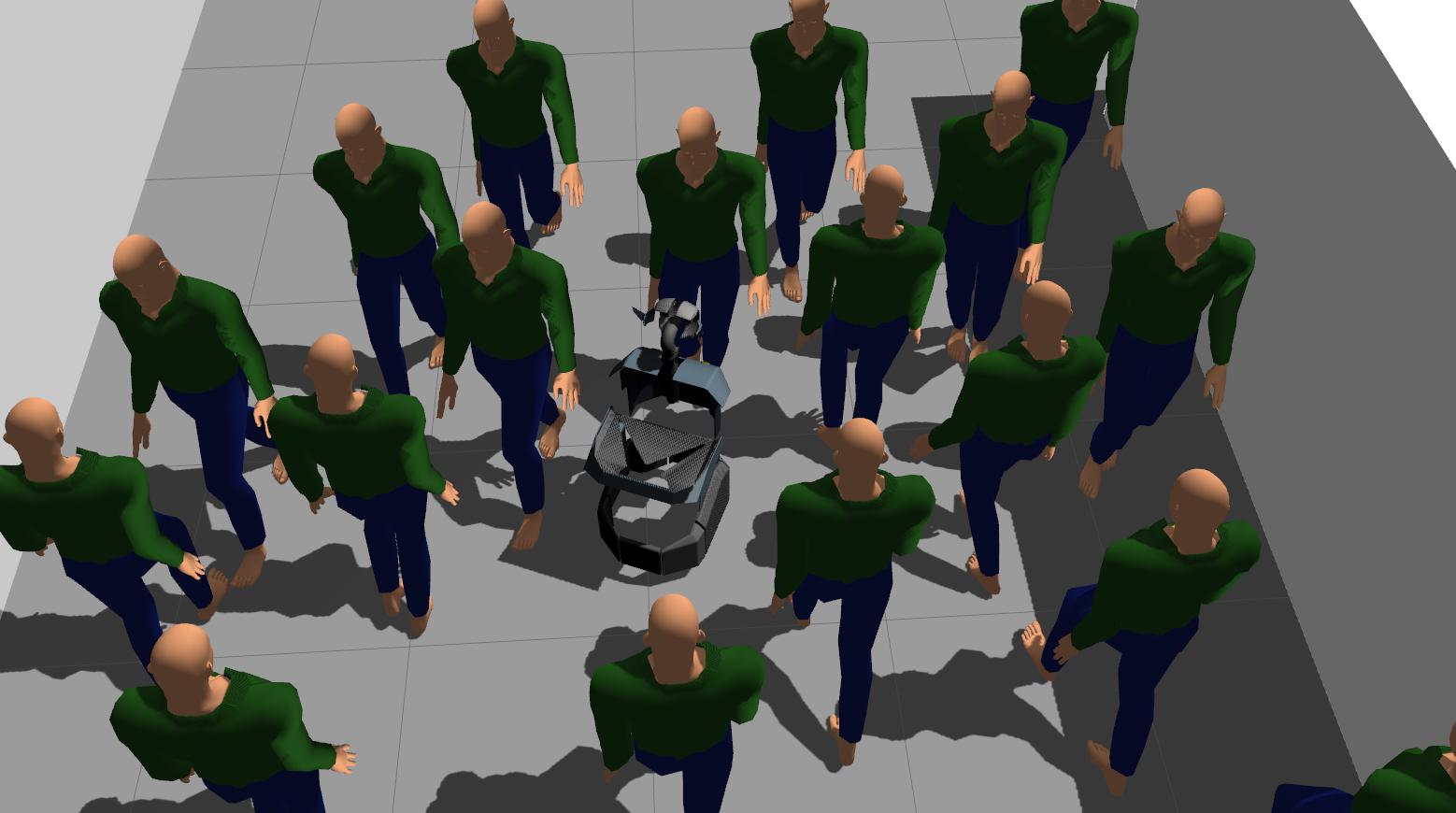}}
     \end{subfigure}
     \begin{subfigure}{0.24\textwidth}
         \centering
         \frame{\includegraphics[width=\textwidth]{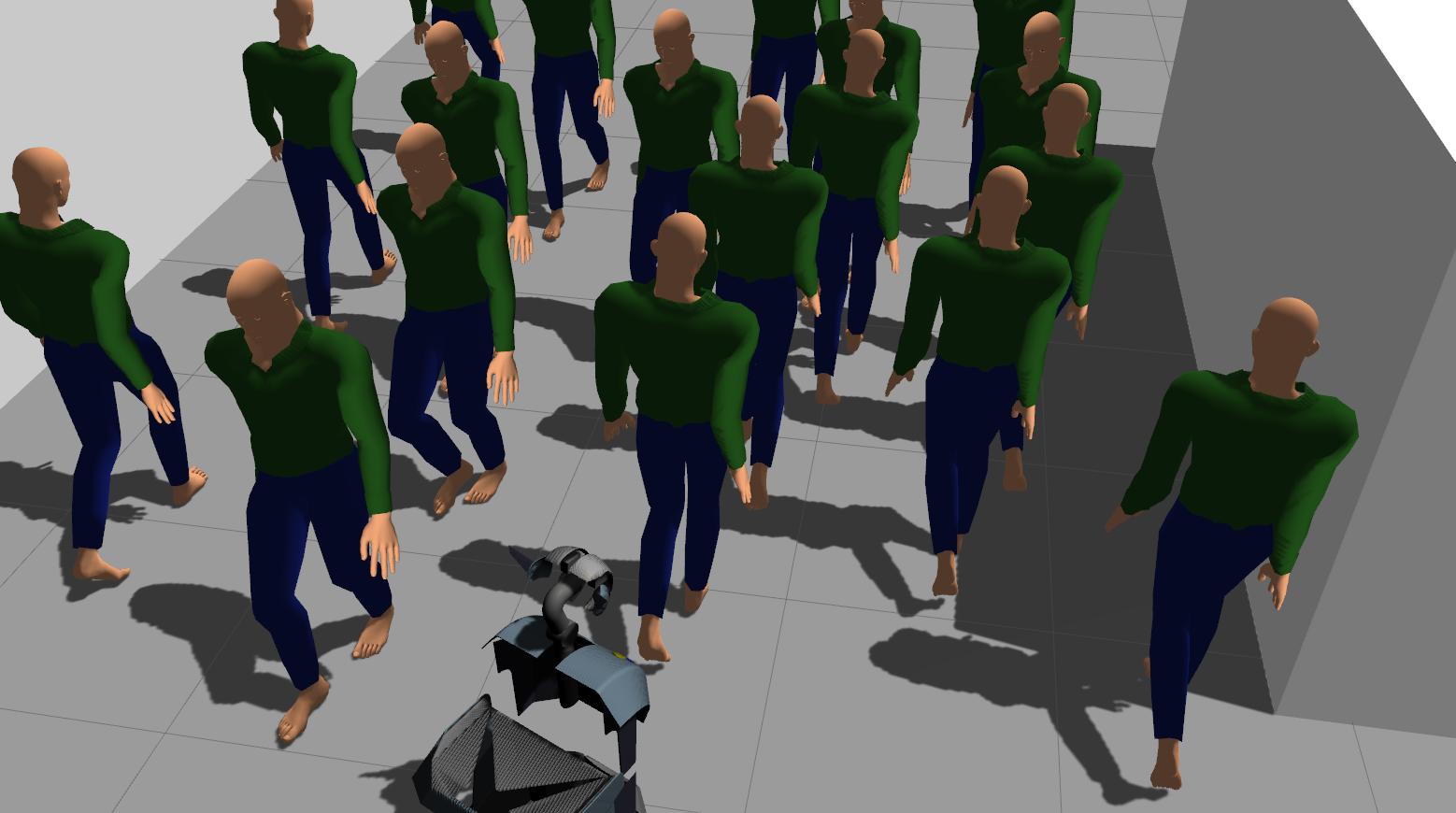}}
     \end{subfigure}
\caption{\emph{Sim 1} setup. (Left): The reactive controller often drives head-on into the crowd, leading the robot to become stuck and resulting in minor collisions. A considerable amount of time is needed to resolve the congestion. (Right): The robot controlled by SPAN evades congestion when possible, and follows behind pedestrians that are moving in the same direction.} \label{CrowdsFig}
\end{figure}
The resulting motion trajectories by the controlled robot in the experiments are illustrated in \cref{expref0}, where the motion trajectories of the robot is marked in red, while those of pedestrians are marked in black. At a glance, we see that the controlled robot navigates smoothly and sensibly. The evaluated metrics are tabulated in \cref{tableresults}. The average computational times of an entire iteration of SPAN, including prediction and obtaining controls, are 27ms, 31ms, 33ms in  \emph{Dataset}, \emph{Sim 1}, \emph{Sim 2} setups respectively, with no iteration taking more than 50ms in any setup.

SPAN is able to control the robot reach the goal, without collisions, in all of the experiment environments. In particular, relative to the compared methods, SPAN demonstrates markedly safer navigation through the \emph{Sim 1} and \emph{Dataset} environment setups. Both of these environments are relatively cluttered, with regions of congestion arising. We observe that both CL-RRT$^*$ and the reactive controller has a greater tendency to freeze amongst large crowds, while SPAN gives more fluid navigation amidst crowds. \Cref{CrowdsFig} shows how the reactive controller (left figure) and SPAN (right figure) navigate the robot through crowds in the dense \emph{Sim 1} environment. The reactive controller often controls the robot to drive towards incoming pedestrians head-on. This behaviour often results in driving into bottleneck positions and congestion arising, with the robot then unable to escape from a trapped position. On the other hand, the robot controlled by SPAN avoids getting stuck in crowds, as the future position of pedestrians are anticipated. Incoming pedestrians are actively evaded, while the robot follows other pedestrians through the crowd. 

\begin{figure}[t]
\centering
\begin{tikzpicture}
\node (img) {
\begin{subfigure}{0.16\textwidth}
         \centering
         \includegraphics[width=\textwidth]{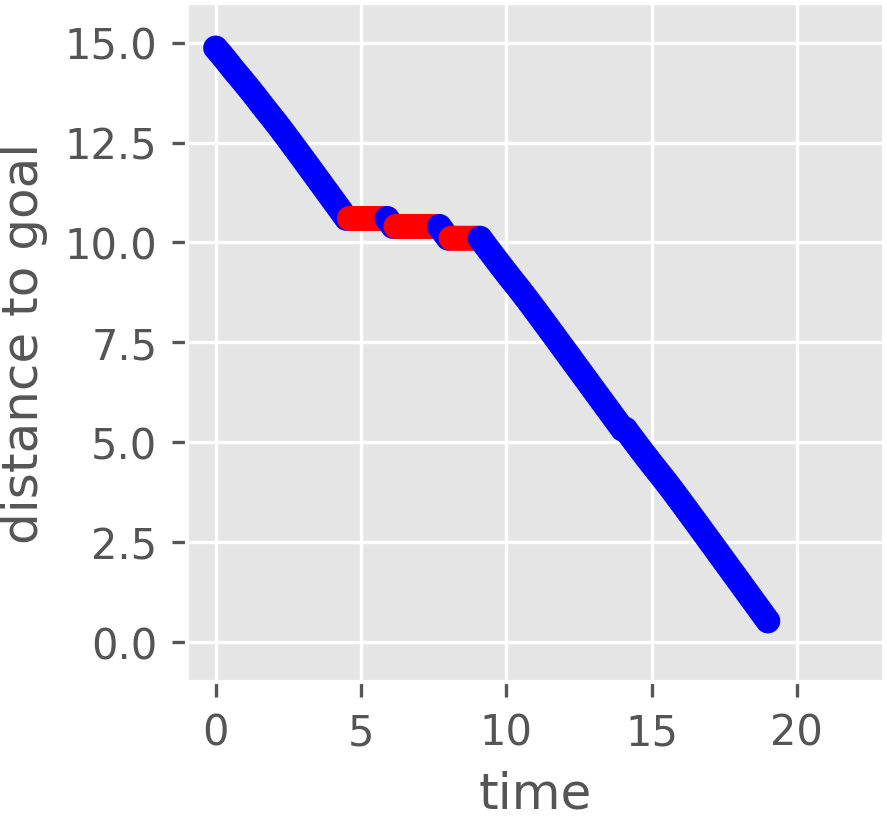}

     \end{subfigure}
     \begin{subfigure}{0.16\textwidth}
         \centering
         \includegraphics[width=\textwidth]{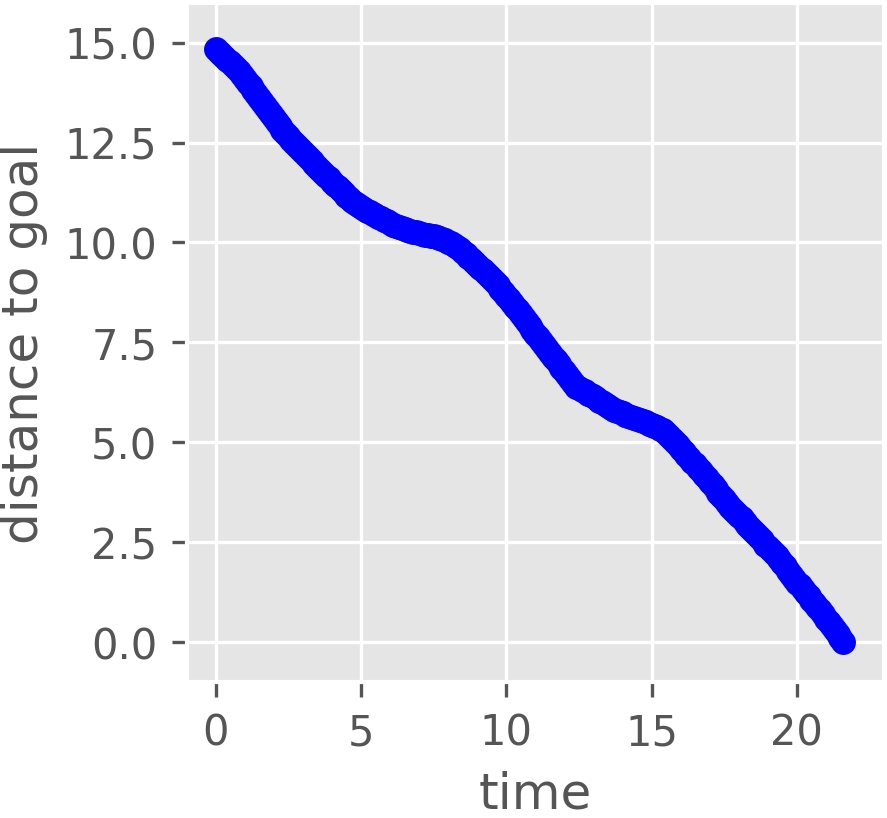}

     \end{subfigure}
     \begin{subfigure}{0.15\textwidth}
         \centering
         \includegraphics[width=\textwidth]{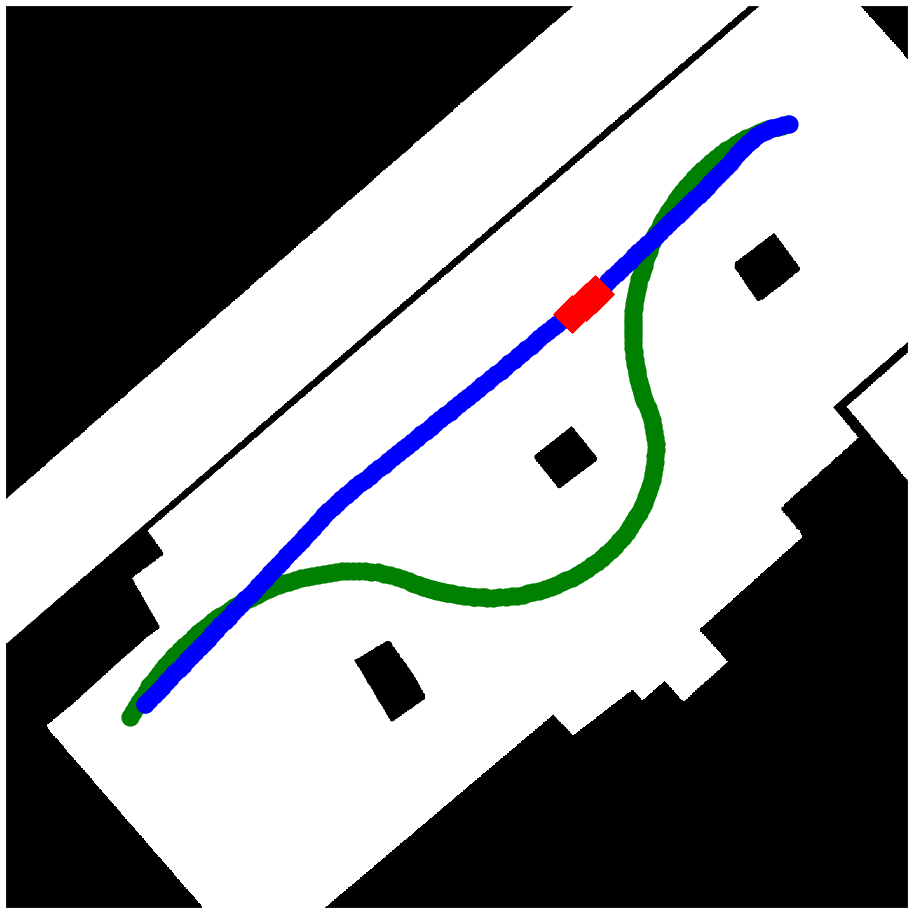}

     \end{subfigure}
     };
    \node at (-3.6,-1.15){\scriptsize \textsc{cl-rrt}$^*$};
    \node at (-0.9,-1.17){\scriptsize \textsc{span}};
\end{tikzpicture}%
     \caption{(Left, Center) shows time against the distance between the robot and goal, for the CL-RRT$^{*}$ and SPAN respectively. The times that are in collision are given in red, while the collision-free times are in blue. (Right) shows the motion of the robot as controlled by CL-RRT$^{*}$ in blue, and that as controlled by ours in green. Time-steps where the robot is in collision are in red. We see that SPAN is able to provide a safe path for the controlled robot, while CL-RRT$^{*}$ provides an aggressive path, resulting in over $3s$ of collision.}\label{aggres}
\end{figure}

In the \emph{Dataset} setup, both the compared CL-RRT$^*$ and the reactive controller produce very collision-prone navigation trajectories. This is due to the completely dominate pedestrian behaviour, which follows collected trajectory data, whereas the simulated pedestrian crowds also attempts to avoid collision with the robot. Both CL-RRT$^*$ and the reactive controller give a relatively straight trajectory towards the goal and are in collision for more than $3s$, colliding with three different pedestrians. Meanwhile SPAN produces a safe and meandering trajectory. The trajectories of the controlled robot by CL-RRT$^{*}$ and SPAN is illustrated in \cref{aggres}, along with plots of how the distance to goal changes in time. We argue, in the \emph{Dataset} setup, although the time-to-goal of SPAN is marginally longer, the navigated path is much safer with no collisions, while the compared methods have more than 3s of collision.  Additionally, all three of the evaluated methods perform similarly in the comparatively open \emph{Sim 2} environment, producing collision-free and efficient  trajectories. This is due to the ample space in the environment, with relative little congestion occurring. 

\begin{figure}[t]
\centering
     \begin{subfigure}{0.15\textwidth}
         \centering
         \frame{\includegraphics[width=\textwidth]{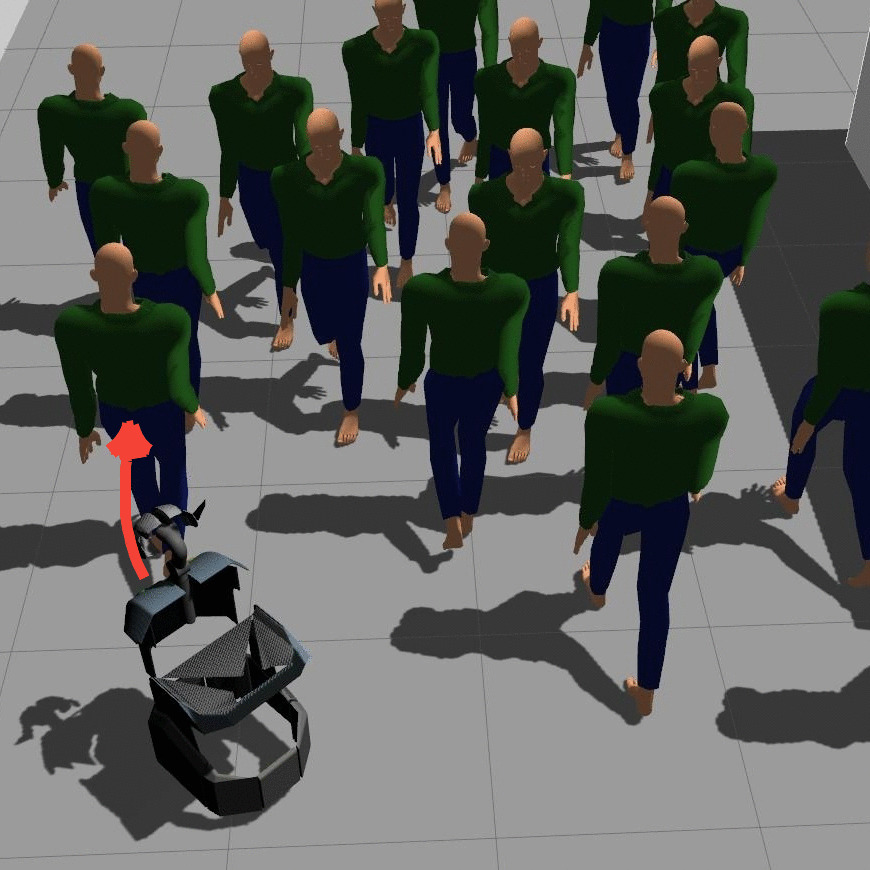}}
     \end{subfigure}
     \begin{subfigure}{0.15\textwidth}
         \centering
         \frame{\includegraphics[width=\textwidth]{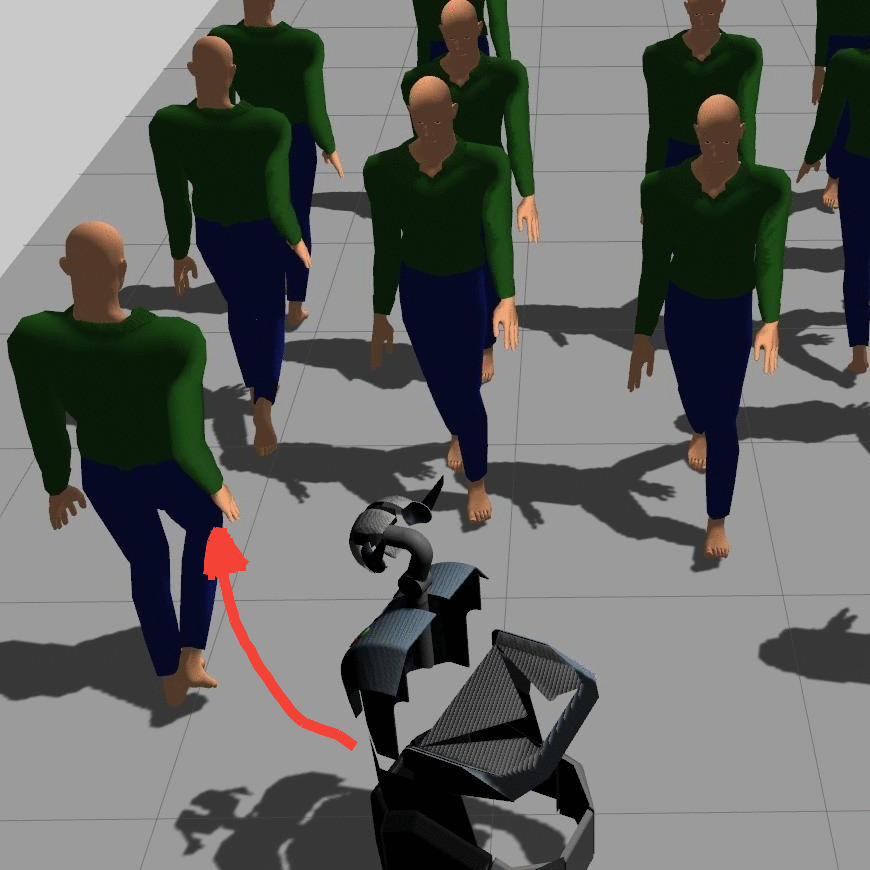}}
     \end{subfigure}
          \begin{subfigure}{0.155\textwidth}
         \centering
         \frame{\includegraphics[width=\textwidth]{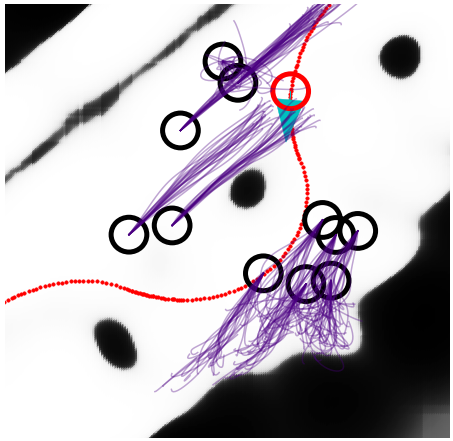}}
     \end{subfigure}
        \caption{Anticipatory behaviour of our controlled robot can be observed. (Left, center) The robot follows (red arrow) pedestrians anticipated to move ahead, while preemptively avoiding incoming pedestrians. (Right) Robot as red circle and pedestrians as black. Sampled trajectories (in purple, next 4.0s) visualise predicted SP, and the robot trajectory is in red. The robot evades an incoming group of pedestrians preemptively, and moves toward pedestrians predicted to move away.}
        \label{expref1}
\end{figure}

\subsection{Emergent Behaviour}

The formulated control problem in \cref{costOpt,costOpta,costOptb,costOptc} does not explicitly optimise for the following of other agents. However, we observe following behaviour, behind other pedestrians, also moving in the goal direction, through crowds. \Cref{expref1} (left, center) show the controlled robot following moving pedestrians through crowds, while avoiding those moving towards the robot. This is also detailed in \cref{expref1} (right), where predicted future positions are visualised by sampling and overlaying trajectories (in purple) from the SP model. The entire navigated motion trajectory taken by the robot is outlined in red. We observe the robot taking a swerve away from the group of incoming pedestrians preemptively, and turning to follow the other group predicted to move away. At that snapshot, the pedestrians in the lower half of the map blocks the path through to the goal, but are all predicted to move away, giving sufficient space for our robot to follow and pass through. The emergence of the crowd-following behaviour facilitates smooth robot navigation through crowds.

\begin{table}[t]
\centering
\scalebox{0.8}{
\begin{tabular}{ll|lll}
\toprule

Method & Metrics               & Dataset & Sim 1 (crowded) & Sim 2 (open) \\ \hline
SPAN (ours)   & TTG       & 21.7        & 25.9        & 17.4    \\
       & DOC & 0.0           & 0.0           & 0.0       \\
CL-RRT$^*$ & TTG        & 19.0          & 36.8        & 17.8    \\
       & DOC & 3.7         & 1.1         & 0.0      \\
Reactive & TTG        &   19.1      & 43.5       & 17.1   \\
Controller & DOC &  3.4      & 0.4        & 0.0      \\
\bottomrule
\end{tabular}}\caption{The evaluations of SPAN against compared methods. TTG gives the time (in seconds) taken by the robot to reach the goal, and DOC gives the total duration (in seconds) the robot is in collision. SPAN finds a collision-free solution in all three evaluated set-ups.}\label{tableresults}

\end{table}
\section{Conclusions and Future Work}
We introduce Stochastic Process Anticipatory Navigation (SPAN), a framework for anticipatory navigation of non-holonomic robots through environments containing crowds and static obstacles. We learn to anticipate the positions of pedestrians, modelling future positions as continuous stochastic processes. Then predictions are used to formulate a time-to-collision control problem. We evaluate SPAN with simulated crowds and real-world pedestrian data. We observe smooth collision-free navigation through challenging environments. A desired emergent behaviour of following behind pedestrians, which move in the same direction, arises. The stochastic process representation of pedestrian motion is compatible as an output for many neural network models, bringing together advances in motion prediction literature with robot navigation. Particularly, conditioning on more environmental factors to produce higher quality predictions within SPAN, is a clear direction for future work.

\bibliographystyle{ieeetr}
\bibliography{biblo}
\end{document}